\pdfoutput=1
\PassOptionsToPackage{table}{xcolor}
\documentclass[11pt]{article}
\usepackage[final]{acl}
\usepackage{times}
\usepackage{graphics}
\usepackage{latexsym}
\usepackage{hyperref}
\usepackage{svg}
\usepackage{amsmath}
\usepackage{multirow}
\usepackage[T1]{fontenc}
\usepackage[utf8]{inputenc}
\usepackage{microtype}

\title{Increasing the Difficulty of Automatically Generated Questions via Reinforcement Learning with Synthetic Preference}

\author{
  William Thorne$\dagger$ \And Ambrose Robinson$\dagger$ \And Bohua Peng$\dagger$ \And Chenghua Lin$\ddagger$ \AND Diana Maynard$\dagger$ \\
  $\dagger$ Department of Computer Science, University of Sheffield \\
  $\ddagger$ Department of Computer Science, University of Manchester \\
  \texttt{\{wthorne1, bpeng10, d.maynard\}@sheffield.ac.uk} \\
  \texttt{ambrose@parablestudio.co.uk} \\
  \texttt{chenghua.lin@manchester.ac.uk}
}

\begin{document}
\maketitle
\begin{abstract}
As the cultural heritage sector increasingly adopts technologies like Retrieval-Augmented Generation (RAG) to provide more personalised search experiences and enable conversations with collections data, the demand for specialised evaluation datasets has grown. While end-to-end system testing is essential, it's equally important to assess individual components. We target the final, answering task, which is well-suited to Machine Reading Comprehension (MRC). Although existing MRC datasets address general domains, they lack the specificity needed for cultural heritage information. Unfortunately, the manual creation of such datasets is prohibitively expensive for most heritage institutions. This paper presents a cost-effective approach for generating domain-specific MRC datasets with increased difficulty using Reinforcement Learning from Human Feedback (RLHF) from synthetic preference data. Our method leverages the performance of existing question-answering models on a subset of SQuAD to create a difficulty metric, assuming that more challenging questions are answered correctly less frequently. This research contributes: (1) A methodology for increasing question difficulty using PPO and synthetic data; (2) Empirical evidence of the method's effectiveness, including human evaluation; (3) An in-depth error analysis and study of emergent phenomena; and (4) An open-source codebase and set of three llama-2-chat adapters for reproducibility and adaptation.

\end{abstract}

\section{Introduction}

The cultural heritage sector is increasingly leveraging advanced technologies like large language models (LLMs) \citep{openaiGPT4TechnicalReport2024, touvronLlamaOpenFoundation2023} and AI assistants \citep{teamGeminiFamilyHighly2023, ClaudeModelFamily2024} to increase and improve access to collections and their associated data. These technologies provide new opportunities for more dynamic and intuitive interactions with heritage content. One particularly promising technology is Retrieval-Augmented Generation (RAG) \citep{lewisRetrievalAugmentedGenerationKnowledgeIntensive2021}, which retrieves relevant information from a database of vectorized content to generate accurate, fact-based responses to user queries. We believe that RAG, and iterations on the approach, will play a significant role in improving the search capabilities of heritage institutions in the coming years.

\begin{figure}
    \centering
    \includegraphics[alt={A comparison of two questions generated for an input prompt from SQuAD. The PPO model is shown to be an increase in difficulty over the supervised-fine-tuned model with respect to range of dependencies, entity disambiguation and not extracting segments from the source text.},width=1\linewidth]{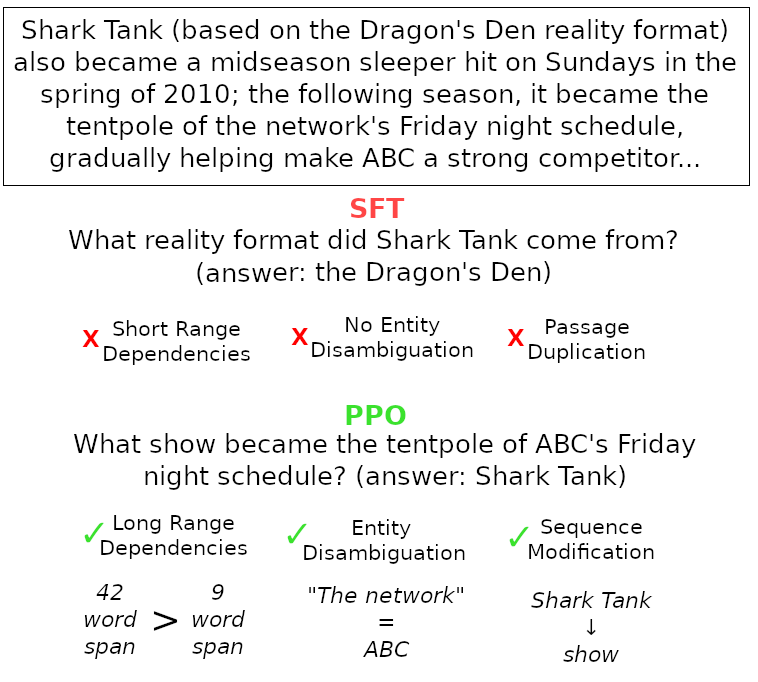}
    \caption{Example generated questions from supervised-fine-tuned question generation model and one fine-tuned with PPO from synthetic difficulty samples.}
    \vspace{-\baselineskip}
    \label{fig:question_comp}
\end{figure}

Heritage search systems are used by the public and academics alike; however, the latter tend to submit more complex and specific queries \citep{koolenInformationRetrievalCultural2009}. RAG has the capability to fulfil these needs but still requires robust evaluation. This includes not only end-to-end system testing but also the evaluation of individual components. As the response is generally required to be written based \textit{only} on the retrieved documents to mitigate language model hallucinations, we argue that the task is one of Machine Reading Comprehension (MRC). While MRC datasets are well-established in the general domain, they are notably lacking in cultural heritage and the cost of their construction is prohibitive for most institutions. We estimate that the popular SQuAD dataset cost about \$12,000 to just write the questions, based on their recommended time per question and stated hourly rate of \$9 \citep{rajpurkar_squad_2016}; the actual cost is likely much higher.

To address these challenges, we propose using Automatic Question Generation (AQG) systems to generate MRC datasets. However, we argue that many automatically generated questions, particularly those from zero- or few-shot approaches, do not provide an adequate challenge for modern language models. Manipulating difficulty is challenging through traditional training approaches given its abstract and subjective nature, and prompt based solutions are intractable when considering the infinite permutations and interactions between different aspects of difficulty \citep{lin_automatically_2015, rajpurkar_squad_2016, beinborn_candidate_2015, hsu_automated_2018,cheng-etal-2021-guiding,alkhuzaey_text-based_2023}.

To control difficulty, we adapt the Reinforcement Learning from Human Feedback protocol used in AI assistant steering \citep{instructgpt, baiTrainingHelpfulHarmless2022}. In this regime, samples are ranked based on specific criteria and paired into \textit{chosen} and \textit{rejected} samples for training a reward model. This reward model learns to distinguish good samples from bad and outputs a signal which steers a policy model. Rather than relying on costly human annotations, we generate synthetic preference data by evaluating question-answering model performance on a subset of SQuAD, assuming that questions answered correctly less frequently are inherently more difficult. This approach leverages the language model's innate feature extraction capabilities, eliminating the need to explicitly define difficulty components. Figure~\ref{fig:question_comp} demonstrates this feature extraction ability by comparing questions generated with and without reinforcement fine-tuning.

We selected SQuAD over an in-domain QA dataset for two main reasons. First, it is a well-studied, large, and diverse dataset. Second, comparable QA datasets at SQuAD's scale are either visual question-answering focused \citep{shengDatasetMultimodalQuestion2016, asprinoLargeVisualQuestion2022} or have data reliability concerns such as OCR text \citep{piryaniChroniclingAmericaQALargescaleQuestion2024}.

This approach enables cultural heritage practitioners to generate challenging evaluation datasets more efficiently and cost-effectively than manual curation. The primary expense is compute resources, which can be accessed in the cloud for only a few dollars per hour.\footnote{\url{https://huggingface.co/pricing}}

We summarise this paper's contributions as follows:

\begin{enumerate}
    \topsep0em
    \itemsep0em
    \item A methodology for increasing the difficulty of automatically generated questions using PPO and synthetic data.
    \item Empirical evidence of the methodology's efficacy including human evaluation.
    \item An in-depth error analysis and study of interesting phenomena that emerge as part of this approach.
    \item An open-source code base and set of models to recreate and adapt our work\footnote{We release all code and a set of three LLaMa-2 adapters on \href{https://github.com/wrmthorne/increasing-AQG-difficulty-via-RLHF}{GitHub}.}.
\end{enumerate}

\section{Related Work}

A similar question generation approach to ours is employed by \citet{zhang_downstream_2022} who adopt a pipeline structure. However, their primary objective is to generate suitable questions rather than specifically focusing on difficulty. An important distinction lies in their extensive pre-processing applied to identify candidate answers before feeding them to the question generation model. We argue that pre-identifying answers may limit diversity and prevent the inclusion of potentially complex answer types.

\noindent\textbf{Analyzing and Controlling Question Difficulty.}~~Understanding and managing question difficulty holds significant importance, especially in tasks involving the creation of exams and assessments \citep{liu2014sherlock,alkhuzaey_text-based_2023}. One approach, as presented by \citet{loginova_towards_2021}, involves modelling the difficulty of multiple-choice questions through the use of softmax scores obtained from a pre-trained QA model. The variance in these scores is then calculated, with higher variance indicating greater difficulty.

\citet{lin2015sherlock} controls the difficulty of quiz questions through the selection of distractor answers based on semantic similarity between linked data items. This involves collecting both structured RDF data and unstructured text, computing similarity scores through K-means clustering, and generating questions and answers via template-based methods. Importantly, the semantic similarity plays a role in determining the difficulty level, with more challenging questions featuring distractors exhibiting higher semantic similarity.

\noindent\textbf{Reinforcement Learning with Human Feedback.}~~RLHF is a machine learning paradigm that combines reinforcement learning with human-provided guidance to steer language models to meet the needs of users, finding frequent use in chatbot and AI assistant settings \citep{instructgpt}. The basis for most modern methods is the Proximal Policy Optimisation (PPO) algorithm \citep{schulman_proximal_2017}, which iteratively enhances the language model's policy to maximize cumulative rewards through interactions with a dataset or language simulation. It collects experiences, evaluates advantages, and updates the policy with a clipped surrogate objective to ensure stability, gradually improving the model's performance. 

\noindent\textbf{Automatic Question Generation.}~~\citet{chen_reinforcement_qa_2019} introduce a cross-entropy loss with a reinforcement learning-based loss function when training a gated bi-directional neural network for question generation. In this context, the reward model is optimising the semantic and syntactic quality of the question. BLEU-4, as a reward function, optimises the model for the evaluation metrics and the negative Word Movers Distance component is used to ensure semantic quality by maximising the similarity between a generated sequence and a ground truth sequence. Although question quality is maintained, other factors such as question difficulty are not considered.

Self-critic sequence training (SCST)~\citep{rennie2017self} uses a classical policy gradient method, REINFORCE, which is a Monte Carlo method. SCST computes rewards with n-gram token overlap as sub-sentence level rewards. Since training sets often have limited questions, these training rewards are arguably sparse, hindering the question generation model from extrapolating beyond the training distribution. 
\citet{liu_generative_qa_2019} adopt a two-component reward for refining ill-formed questions. Question wording is used as a measure of short-term reward, and alignment between the question and answer represents a long-term component. 

\section{Method}

To challenge the high cost of manual annotation while maintaining quality and increasing difficulty, we design and implement a robust system capable of generating contextually relevant, coherent, and challenging question-answer pairs from textual input. The process follows the core methodology of RLHF, deviating only in the use of synthetic preference data to train a reward model. Rather than explicitly defining the characteristics of difficulty and risking failure to capture certain aspects, we exploit the ability of leading question-answer models to derive which questions are challenging, and allow a reward model to extract the component features of the task.

We task three QA models with answering all questions in our validation split of SQuAD. These questions are assigned a score based on the number of times they were answered incorrectly, which are in turn used to generate pairwise preference data. These pairwise samples enable the training of a reward model (RM) for use in fine-tuning a supervised model (SFT) on the task of question generation using Proximal Policy Optimisation (PPO)\citep{schulman_proximal_2017}.

\begin{figure*}
    \centering
    \includegraphics[alt={A graphical depiction of the project pipeline},width=1\linewidth]{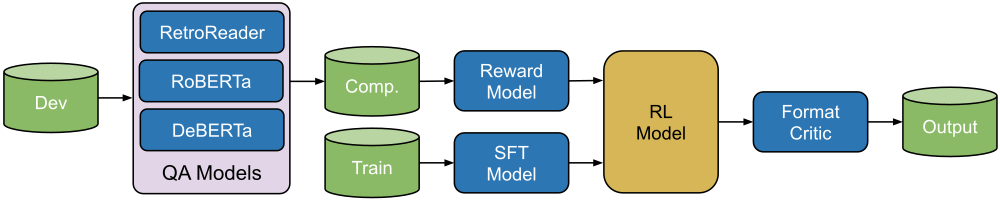}
    \caption{Depiction of our dataset generation pipeline. Question-Answering models are first used to create pairwise comparison data to train a reward model. An SFT model is trained on the train split of SQuAD and then fine-tuned using the reward model, producing the RL model. When generating question-answer pairs for the final dataset, generations are passed through the format critics to ensure data quality.}
    \label{fig:project pipeline}
\end{figure*}

We embed this synthetic RLHF process into a greater pipeline for generating samples, shown in Figure~\ref{fig:project pipeline}. This ensures the quality of the final dataset. The pipeline also contains a set of rule-based critics which are used to exclude samples that are malformed and those with non-unique answers in the source text. Samples are then deduplicated using exact string matching.

The remainder of this section discusses each of the relevant components of the pipeline and the RLHF process. 

\subsection{Supervised Fine-Tuning} \label{sft}

In our training process for question generation and response formatting, we begin by employing a reformatted version of the SQuAD v1 training split (see Table~\ref{tab:squad_splits}). The reformatting converts SQuAD to the task of question-answer pair generation, as shown in Figure~\ref{figure:inverted_squad}. We select the "correct" answer as the one that appears most frequently in the list of answers for each question in the dataset, selecting randomly among the most common if there is no victor. To ensure model robustness without overfitting, the model undergoes a single epoch of training, enabling it to effectively capture the nuances of the task.

\begin{figure}[]
\noindent\fbox{%
    \parbox{\linewidth}{
        \paragraph{Instruction} Write 1 answerable span extraction question and provide the correct answer based on the text. 
        \paragraph{Input} ... Upon its arrival in Canberra, the Olympic flame was presented by Chinese officials to local Aboriginal elder \underline{Agnes Shea}, of the Ngunnawal people. She, in turn, offered them a message stick ...
        \paragraph{Response}  Who received the flame from Chinese officials in Canberra? (answer: \underline{Agnes Shea})
    }
}
\caption{Example training sample from the reformatted SQuAD dataset for use in supervised fine-tuning.} \label{figure:inverted_squad}
\vspace{-\baselineskip}
\end{figure}

\subsection{Reward Modelling}

To control the difficulty of our generated questions, we leverage the intrinsic properties present in challenging questions from SQuAD. To extract these attributes, we employ three question answering models that almost match or exceed human performance on SQuAD v2 to evaluate our development split: a RoBERTa-large model\footnote{\href{https://huggingface.co/deepset/roberta-large-squad2}{deepset/roberta-large-squad2}}, a DeBERTa-large model\footnote{\href{https://huggingface.co/deepset/deberta-v3-large-squad2}{deepset/deberta-v3-large-squad2}} and RetroReader \citep{retro_reader_2020}. Each question is assigned a score based on the number of models that failed to correctly answer the question. These scores are used to place questions into a pairwise ranking setup against other questions for the same input context. Where a question's scores are equal, they are considered ties, and no pairwise sample is created. We also record the margin, defined as the difference in score between the chosen and rejected samples, to experiment with the marginal ranking loss, as defined in \citet{touvron_llama2_2023}. 

\subsubsection{Format Critics}

To ensure the quality of the final dataset, we utilise a collection of rule-based critics which we call \textit{Format Critics}. These critics have two main functions: they remove questions that don't adhere to the desired format of \textit{Q? (answer: A)}; they ensure the provided answer is unique in the text, minimising the number of ambiguous or impossible questions. Samples that pass these critics are then deduplicated using exact matching.

\subsection{Reinforcement Training}

We use Proximal Policy Optimisation \citep{schulman_proximal_2017} with multiple sets of adapters to reduce the memory overhead during training, implemented using the Transformers Reinforcement Learning library \citep{vonwerra2022trl}. A single base model is used with separate LoRA adapters for the policy, reference, and reward model components; each is switched to perform the relevant aspect of the reinforcement training process.

During early experiments, we found that training was often very unstable or resulted in low pass rates at the format critic. To combat this, we added a rule-based reward component to penalise generations that did not pass the format critic. This simple function converts the reward to be the negative absolute reward in the case that samples are malformed. Using a rule-based reward that manipulates the original reward prevents the instability caused by hard coding a fixed penalty and saves the computational complexity and imperfection of a second adapter-based reward model:
\begin{equation}
R_i =
\begin{cases}
  -|R_i| & \text{if malformed} \\
  R_i & \text{otherwise}
\end{cases}
\end{equation}

\section{Experimental Setup}

\subsection{Models}

We conduct our experiments with LLaMa2-7B-chat and apply LoRA adapters to all linear layers for all models. This drastically lowers the number of tunable parameters over full-finetuning, enabling training on a single A100 80GB GPU. We also make use of Flash Attention 2 \citep{dao_flashattention-2_2023} to improve computational and memory efficiency. All LoRA adapters share the same hyperparameters: a LoRA rank of 16, as \citet{dettmers_qlora_2023} found rank to have minimal impact on task performance while enabling larger batches through reduced memory usage. This memory efficiency further allowed us to implement sample packing, particularly beneficial with Flash Attention 2's preference for minimal padding. We set alpha to twice the rank \footnote{\url{https://lightning.ai/pages/community/lora-insights/}}, use a dropout of 0.05  - shown optimal for 7B models by \citet{dettmers_qlora_2023}, and maintain LLaMa-2's BF16. As a baseline, we compare to LLaMa-2-7B-chat in a zero-shot setting (see Appendix~\ref{app:zero_shot}).

We experiment with marginal ranking loss to help distinguish between slight and significant differences in question difficulty while training the reward model. Under the hypothesis that the difficulty of a question is not independent of the associated passage of text, we also experiment with training a reward model with and without the input text attached. Results of these experiments can be found in Appendix~\ref{app:rm_performance}.

\subsection{Generation Settings}

During generation, the model is tasked with producing a single output for each question in the training set using nucleus sampling \citep{holtzman_curious_2020}. We maintain the original configuration for LLaMa-2 with a repetition penalty of 1.1, top P of 0.7, and top K of 0 but increase the temperature from 0.6 to 0.9 to increase the diversity of generations.

\subsection{Data Splits}

We base our splits off the original SQuAD to minimise the risk of data leakage. We maintain the full train split unchanged as any model previously trained on SQuAD will have seen the full train split. We extract a test split of 500 contexts from the dev split, ensuring no contexts appear in both the dev and test splits. We extract 50 unique contexts from the test split for a human evaluation of question quality and answerability. In all cases, context-question pairs were only kept if they fit into the context length of LLaMa-2 when formatted in the correct prompt format. All samples were formatted into the three instruction components: \textit{instruction}, \textit{input}, \textit{response} as shown in Figure~\ref{figure:inverted_squad}.

Only the dev set of our SQuAD dataset was used to derive difficulty comparison data, to ensure the reward model never sees the samples used for evaluation. To evaluate the reward model, we extract 10\% of the comparison contexts. Full dataset statistics can be found in Table~\ref{tab:squad_splits}.

\begin{table}
    \centering
    \begin{tabular}{ccc}
        \hline  
        \textbf{Split} & \textbf{\# Contexts} & \textbf{\# Questions} \\
        \hline
        Train & 18,891 & 87,599 \\
        Dev & 1,567 & 8,038 \\
        Test & 500 & 2,532 \\
        Human Test & 50 & 50 \\
        \hline
        Train comp. & 1,107 & 8,394 \\
        Dev comp. & 123 & 950 \\
        \hline
    \end{tabular}
    \caption{Split of contexts and questions from SQuAD. The \textit{comp.} splits are derived from the dev split, used to evaluate the performance of the reward model during training.}
    \label{tab:squad_splits}
    \vspace{-\baselineskip}
\end{table}

\subsection{Evaluation Metrics}

As our goal is to evaluate the difficulty of answerable questions, we provide the input passage, question and answer to GPT-4o\footnote{gpt-4o as of 1st June 2024} and Gemini-1.5-pro\footnote{gemini-1.5-pro as of 1st June 2024} and ask whether the sample meets our specification of validity. We take samples to be answerable if they were unanimously labelled as such, and reject all other samples. GPT-based evaluations have demonstrated a robust alignment with human preferences across various complex tasks in reference-free settings \citep{fu_gptscore_2023, liu_g-eval_2023}. The results of this analysis can be found in Appendix~\ref{app:llm_ans}.

\begin{table*}
    \centering
    \begin{tabular}{l|c|c|c|c}
        \textbf{Model} & \textbf{Total Valid ($\uparrow$)} & \textbf{DeBERTa ($\downarrow$)} & \textbf{RoBERTa ($\downarrow$)} & \textbf{RetroReader ($\downarrow$)} \\
        \hline
        \textbf{SQuAD} & 2,532 (-) & 0.68 & 0.68 & 0.65 \\
        \hline
        \textbf{ZeroShot} & 357 $\pm$ 14 (0.14) & $0.644 \pm 0.007$ & $0.650 \pm 0.007$ & $0.629 \pm 0.009$ \\
        \textbf{SFT} & 1252 $\pm$ 2 (0.49) & $0.654 \pm 0.012$ & $0.653 \pm 0.005$ & $0.616 \pm 0.015$ \\
        \textbf{PPO-input} & \textbf{1375 $\pm$ 18 (0.54)} & \textbf{0.601 $\pm$ 0.004} & \textbf{0.606 $\pm$ 0.003} & \textbf{0.582 $\pm$ 0.007} \\
        \textbf{PPO-input-margin} & 1373 $\pm$ 4 (0.54) & $0.612 \pm 0.001$ & $0.608 \pm 0.005$ & $0.587 \pm 0.002$ \\
    \end{tabular}
    \caption{Question-Answering model performance on each set of samples. Models were only supplied samples which passed the format critics and were unanimously deemed answerable by GPT-4o and Gemini-1.5-pro. The \textit{Total Valid} column indicates this number of valid samples used during question answering. Accuracy is based on exact match and results are mean and standard deviation across three sets of generated samples. Lower accuracy indicates harder questions.}
    \label{tab:main_results}
\end{table*}

To assess the quality of generated questions relative to our SQuAD test split, we \textit{intentionally avoid} $n$-gram based metrics such as BLEU \citep{papineni_bleu_2002}, ROUGE \citep{lin_rouge_2004}, and more modern alternatives such as Q-Metrics \citep{nema_towards_2018}, as we believe they restrict diversity of generation, constraining the model to reference questions and answers. We instead adopt the following reference-free metrics:

\textbf{Syntactic Divergence} provides a distance measure between two dependency paths which acts as a measure of difficulty. Word-lemma anchors, common to both the question and answer sentence, are first detected. A dependency path from the anchor to the interrogative word (who, what, etc.) in the question is compared to the dependency path between the anchor and the answer span in the answer sentence using Levenshtein distance \citep{levenshtein1966binary}.

\textbf{RQUGE} calculates an \textit{acceptability-score} by generating an answer for the candidate question and predicting the semantic similarity between the predicted answer and the gold answer provided by the user. In our setup, this metric acts as an assessment of both the question and answer quality \citep{mohammadshahi_rquge_2023}.

\textbf{QAScore} attempts to align AQG evaluation to human judgements. Question-answer pairs are evaluated by summing log-probabilities of RoBERTa correct token predictions for all words in the answer when masked individually. QAScore claims to show strong correlation with human judgement (Spearman $r = 0.864$) \citep{ji_qascoreunsupervised_2022}.

\textbf{Self-BLEU} assesses how similar questions are to other questions generated for a given context. Each question is taken as a hypothesis and the others as a reference for the BLEU calculation. The self-BLEU is taken as the average BLEU for the question collection \citep{zhu_texygen_2018}. 

\section{Results and Discussion}

\textbf{Model Accuracy.}~~To measure performance, we observe the difference in prediction accuracy for QA models on each dataset. Table~\ref{tab:main_results} shows that in all cases of PPO training, we observe a decrease in average model prediction accuracy and an increase in the total number of valid generations. The consistent decrease in absolute prediction accuracy for all models when using the PPO trained models over both zero-shot and SFT signifies an increase in average question difficulty. The SFT process vastly improves the model's ability to generate valid questions. The PPO process further bolsters this capability which illustrates that the model is learning the intrinsic properties of high-quality questions. The performance of the reward models, shown in Appendix~\ref{app:rm_performance}, is reflected here, showing lesser degrees of improvement for those models fine-tuned without access to the input passage. 

\begin{figure*}
    \centering
    \includegraphics[alt={KDE Plots showing the distribution of scores for each metric, for each model. ZeroShot is shown to have anomalous syntactic divergence, SQuAD performs significantly better on RQUGE, Most models are matched on QA-Score and ZeroShot once again shows an anomalously high diversity on self-bleu.},width=1\linewidth]{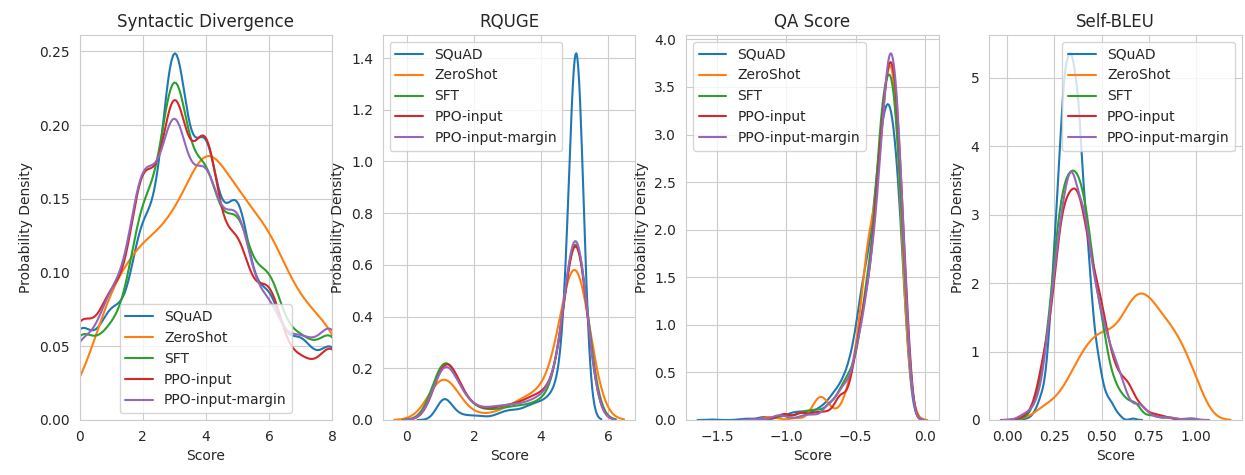}
    \caption{Distribution of reference free metrics results for each model's generations based on our SQuAD test set.}
    \label{fig:ref_free_results}
\end{figure*}

\noindent\textbf{External Metrics.}~~ Figure~\ref{fig:ref_free_results} shows results for the reference-free metrics. RQUGE is clearly effective at discriminating between human-written SQuAD samples, those generated by the fine-tuned models and the zero-shot examples, but it is unable to separate out the SFT and PPO results. The particularly high score for SQuAD could in part be due to data leakage as the answer generation model for the metric was trained on SQuAD \citep{khashabi_unifiedqa-v2_2022}. This would indicate why our newly generated questions might score lower as it cannot have memorised the answer. Syntactic divergence results for the SQuAD test split and all trained model generations follow a consistent distribution but the zero-shot results appear much better, despite having a higher average prediction accuracy than the SFT and PPO models. Zero-shot obtaining higher syntactic divergence could stem from the general purpose language generation objective of LLaMa-2-chat. This could cause the model to generate boilerplate text which distances the structure of the question from that of the answer sentence but doesn't necessarily result in a more difficult question. QAScore proves uninformative, only being able to subtly identify SQuAD samples from model generated samples. Self-BLEU indicates that SQuAD samples are the most diverse, which is to be expected, but that zero-shot samples exhibit a distinct lack of diversity when compared with fine-tuned models. This result is, in part, misleading as Self-BLEU was only calculable for input passages with at least two valid questions. As the number of valid generations was so low for the zero-shot model, the cases where multiple valid questions were generated for a context was disproportionately in favour of identical generations.

In general we find the reference-free metrics to show limited correlation with model prediction accuracy and an ability to differentiate human written samples from model generations. We believe this is evidence for the continued need for more reliable, reference-free evaluation tools for question generation.

\noindent\textbf{Human Evaluation.}~~To evaluate question quality, we conduct a human evaluation on a subset of 50 passages from the test split. Each input passage and question is filtered through the format critic then provided to two annotators who select either the correct answer span or indicate that the question cannot be answered. In the case of annotator disagreement or the annotated answers differing from the model generated answer, the annotator responses and the model answer are provided to two new annotators who both select which responses are appropriate. We allow annotators to select multiple responses as correct but only include those that were selected unanimously by both annotators as valid. We observe an agreement of $\kappa = 0.7975$ between annotators. The results of this evaluation, shown in Table~\ref{tab:human_eval_results}, displays an equivalent or improved rate of answerability when fine-tuning with PPO; the answerability proportions for each dataset are roughly equivalent to those presented in Table~\ref{tab:main_results}. This further corroborates the efficacy of our approach.

\begin{table}
    \centering
    \begin{tabular}{l|c|c}
        \textbf{Model} & \textbf{Full} & \textbf{Partial} \\
        \hline
        \textbf{ZeroShot} & 0.10 & 0.14 \\
        \textbf{SFT} & 0.52 & 0.60 \\
        \textbf{PPO-input} & 0.52 & 0.64 \\
        \textbf{PPO-input-margin} & \textbf{0.56} & \textbf{0.64} \\
    \end{tabular}
    \caption{Results of human evaluation for question quality. \textit{Full} indicates that the model generated answer was a valid answer according to the format critics and identified by human annotators and \textit{Partial} indicates that the sample passed format critics and a valid answer was identified for the question but the model generated answer did not match.}
    \label{tab:human_eval_results}
    \vspace{-\baselineskip}
\end{table}

The results demonstrate that reinforcement learning can effectively manipulate question difficulty, while highlighting important avenues for future work. While SQuAD's synchronic nature served our experimental needs, cultural heritage datasets typically present diachronic challenges that add complexity to question generation. Although specialised diachronic models exist \citet{drinkallTimeMachineGPT2024}, they lack the extensive training of general-domain LLMs. However, these larger models' exposure to historical corpora, combined with their advanced instruction-following capabilities, suggests potential for manipulating temporal complexity as an additional dimension of question difficulty.

\subsection{Error Analysis}

\begin{figure*}
    \centering
    \includegraphics[alt={Four pie charts, ordered left to right as PPO-input-margin, SFT, ZeroShot, and PPO-in},width=1\linewidth]{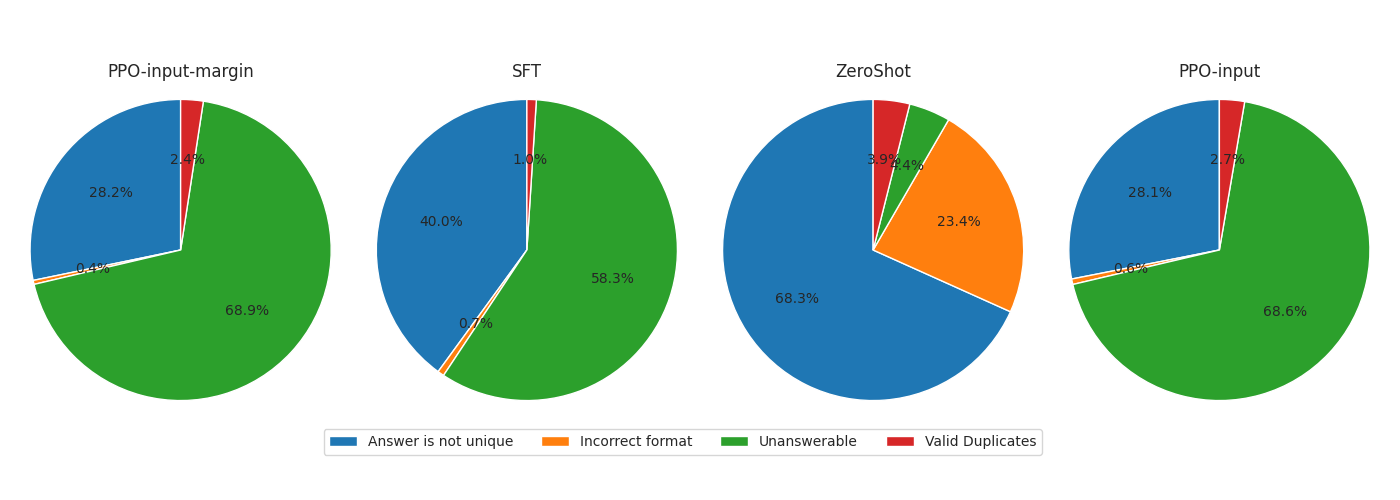}
    \caption{Error distribution of questions for SFT, ZeroShot, and the two best performing PPO variants. }
    \label{fig:error-dist}
\end{figure*}

\textbf{Failure Modes.}~~At a high level, we can observe the reasons for sample rejection for each model. As shown in Figure~\ref{fig:error-dist}, the zero-shot model is generally unable to generate samples that have a single answer span in the text, despite exactly specifying this in the prompt. The high number of incorrectly formatted samples was a result of only a question being generated or neither a question nor answer being generated. For all the trained model variants, the dominant failure mode was unanswerable questions. As shown in Appendix~\ref{app:llm_ans}, each of the fine-tuned models show a similar proportion of otherwise valid samples being unanswerable. The answerability rate could potentially be improved by generating candidate answers, as in \citep{zhang_downstream_2022}, and passing an input passage and answer to the question generation model.

\begin{figure}
    \centering
    \includegraphics[alt={A plot showing the distribution of model generated answers about which it also generated the question. All models show a very strong bias to the first 20\% with the exception of ZeroShot whos peak is at around 22-25\%},width=1\linewidth]{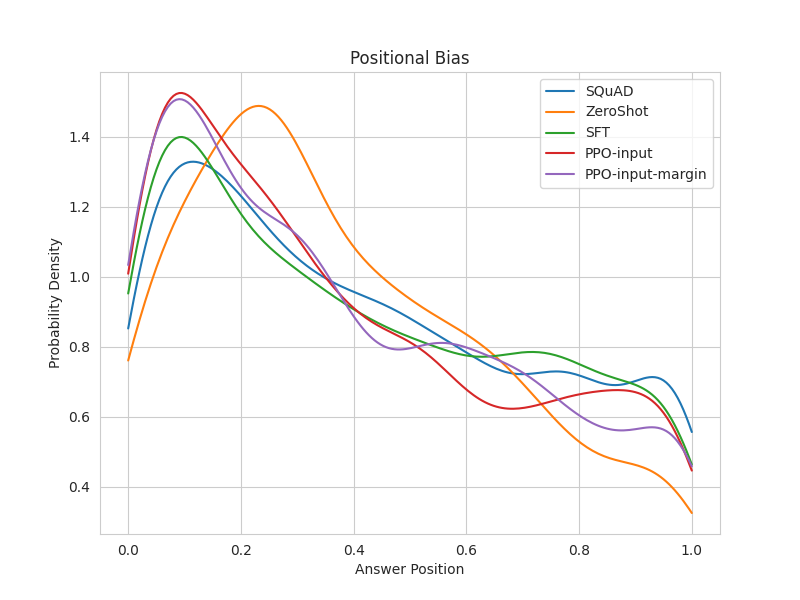}
    \caption{Position of answer span, merged to be a single word, as a proportion of the way through the input passage when split into words. SQuAD positions are selected from our test split and answers are chosen to be the most common from the list of suitable answers. Neither invalid nor exact duplicate questions are considered.}
		\vspace{-\baselineskip}
    \label{fig:position-bias}
\end{figure}

\noindent\textbf{Positional Bias.}~~One interesting phenomenon is the positional bias in where the model chooses to generate answers. To calculate positional bias, we treat the full answer span as a single "word" and calculate the proportion through the input paragraph in which the answer word appears. As seen in Figure~\ref{fig:position-bias}, the zero-shot positional bias is less severe than in the other datasets. The positional bias of SQuAD is clearly seen as, after training on the dataset, all models exhibit this same preference for the beginning of input passages. The clear bias observed in the zero-shot model, despite not being fine-tuned, is documented in other tasks such as LLM ranking \citep{wang_large_2023, li_split_2023} and in summarisation where introductory content is favoured \citep{ravaut_context_2023}. A potential remedy is to supply the model with a sliding window of sentences across the context paragraph to force the model to generate questions throughout the text. While this would improve the diversity of a final dataset, it may have the adverse effect of limiting the range of dependencies, restricting potentially challenging questions across the whole text. 

\noindent\textbf{Hallucinated External Knowledge.}~~Where ambiguous references to specific entities exist in the input passage such as \emph{the museum collection}, the models frequently attempt to fill in which entity is being referred to. From a context containing ambiguous references to an unnamed museum, the questions \emph{What year did the Tate acquire the statue of St John the Baptist?}, \emph{How many works does Rodin have in the British Museum's collection?} were generated across both the SFT and PPO models; the examples consistently passed LLM evaluations of answerability. This suggests the solution to this problem is more holistic and requires improvements at a foundational model level to resolve. We could resolve this at a critic level through more careful prompting, however, this returns to our original and intractable task of textually describing a complex task. A more holistic solution could be to adapt PPO with functional grounding \citep{carta2023grounding} to be a pure text task. However, this may lower the quality of questions as it could discourage the use of implicit or complementary knowledge.

\noindent\textbf{Unidirectional Relationships.}~~A strategy to increase the difficulty of questions is to invert relationships found in the text. The models sometimes misappropriate this tool, resulting in invalid questions such as the question \emph{What did the Ming dynasty represent?} from a passage containing \emph{...explorer Zheng He representing the Ming Dynasty...}. Knowledge graph assisted generation could help to resolve these logical inconsistencies \citep{lin_automatically_2015}. However, expecting our target demographics, emerging domains, to possess high-quality knowledge graphs is an unreasonable assumption.

\section{Conclusion}
In this paper, we introduce a low-cost methodology for generating challenging MRC datasets to meet the growing need for evaluation datasets in the cultural heritage sector. By using high-performing question-answering models to identify the most difficult questions, we were able to create synthetic pairwise data for training a reward model. Rather than manually defining question difficulty, our approach allows the model to learn and extract these features autonomously, leading to a significant improvement in the difficulty of questions generated for evaluation.

With this said, we trained on a general domain dataset in order to single out the training behaviour, in doing so losing many of the characteristic features of heritage datasets. In future work we will examine how the training paradigm fares under the unique challenges presented by such a varied industry.

Although this work was produced to meet the evaluation demands of our ongoing work in RAG at our institution, we also highlight that the approach can work in any domain and that with some modification, it could be used to augment other dataset formats. 
We believe this approach can be extended further, allowing for the manipulation of multiple abstract properties simultaneously through multi-reward model setups \citep{wu_fine-grained_2023}.

\section*{Limitations}

This project only shows the suitability of the method on a single model. In future work, we seek to address this by performing a more comprehensive review of the approach across a range of model sizes and architectures. We also acknowledge that this method currently only addresses answerable questions while most contemporary QA datasets utilise both answerable and unanswerable questions. Finally, despite using LoRA and multi-adapter training, we still required approximately 15 GPU hours on an A100 80GB which restricts the potential audience for this approach. Evaluating smaller models or quantisation will enable greater access to this project's benefits.

\section*{Acknowledgements}

This work was supported by the Arts and Humanities Research Council [grant number AH/X004201/1].

\section*{Ethics Statement}

This project has been approved by the relevant institution's ethics committee. We use LLaMa2 in accordance with Meta's license\footnote{\url{https://ai.meta.com/llama/license/}}. All annotators were located through word of mouth and paid £12 per hour - above the UK National Living Wage of £11.44.

\bibliography{anthology,custom}

\appendix

\section{Reward Model Performance} \label{app:rm_performance}

To understand the relative contributions of marginal ranking loss and the use of the input when training reward models to discriminate based on difficulty, we trained all four permutations of settings on the whole training split of the comparisons dataset and evaluated on the test split. As shown in Table~\ref{tab:rm_results}, the inclusion of the input text had a very significant impact on performance. This was expected as the difficulty of a question is not independent of the related passage. Surprisingly, marginal ranking loss had a very slight negative impact on reward model performance. We believe this could be due to the fact that features of difficulty are very subtle and the marginal component may have caused too significant adjustments due to higher loss values. 

\begin{table}
    \centering
    \begin{tabular}{l|c}
        \textbf{Model} & \textbf{Accuracy (\%)} \\
        \hline
        \textbf{RM} & 63.66 \\
        \textbf{RM-input} & \textbf{70.69} \\
        \textbf{RM-margin} & 62.39 \\
        \textbf{RM-input-margin} & 70.38 \\
    \end{tabular}
    \caption{Accuracy of reward model variants based on the test split of the comparisons dataset. \textit{input} indicates that the model was trained with the question and associated text passage as input and \textit{margin} indicates that marginal ranking loss was used.}
    \label{tab:rm_results}
\end{table}

\begin{figure}[hbp]
\noindent\fbox{%
    \parbox{\linewidth}{
        Following is a text, a question and an answer. You must determine whether the provided answer is a correct span-extraction response to the question. If there are multiple plausible answers in the text, the answer should be the most relevant or accurate one. If there are multiple equally plausible answers in the text, respond "NO". If the provided answer is incomplete or contains excess information, respond "NO". If the answer does not correctly answer the question, respond "NO". Only if the answer is correct and does not breach the aforementioned requirements, respond with "YES". \\
        \textbf{Text}: ... Upon its arrival in Canberra, the Olympic flame was presented by Chinese officials to local Aboriginal elder Agnes Shea, of the Ngunnawal people. She, in turn, offered them a message stick ... \\
        \textbf{Question}:  Who received the flame from Chinese officials in Canberra? \\
        \textbf{Answer}: Agnes Shea \\

        Respond with only "YES" or "NO" in response to this task. Do NOT provide any other text or reasoning.
    }
}
\caption{Example prompt and response to GPT-4o (gpt-4o as of 1st June 2024) and Gemini-1.5-pro (gemini-1.5-pro as of 1st June 2024).} \label{figure:gpt4_prompt}
\end{figure}

\begin{table*}
    \centering
    \begin{tabular}{l|c|c|c|c}
        \textbf{Model} & \textbf{Answerable ($\uparrow$)} & \textbf{Unanswerable ($\downarrow$)} & \textbf{Undetermined ($\downarrow$)} & \textbf{Cohen's $\kappa$ ($\uparrow$)} \\
        \hline
        \textbf{ZeroShot} & \textbf{0.73} & \textbf{0.14} & \textbf{0.13} & 0.61 \\
        \textbf{SFT} & 0.64 & 0.20 & 0.16 & \textbf{0.62} \\
        \textbf{PPO} & 0.64 & 0.20 & 0.16 & \textbf{0.62} \\
        \textbf{PPO-input} & 0.62 & 0.20 & 0.18 & 0.58 \\
        \textbf{PPO-margin} & 0.62 & 0.19 & 0.19 & 0.56 \\
        \textbf{PPO-input-margin} & 0.63 & 0.21 & 0.16 & \textbf{0.62} \\
    \end{tabular}
    \caption{Results of answerability task posed to GPT-4o and Gemini-1.5-pro. Results represent the proportion of samples that are answerable, unanswerable and undecided, taken from those samples which passed the format critic.}
    \label{tab:ans_results}
\end{table*}

\section{Obtaining Zero-Shot Model Generations} \label{app:zero_shot}

To obtain zero-shot generations, we adopt a slightly different approach. To avoid overconstaining the output of the model, we adopted a two-stage process. LLaMa-2-7b-chat was first tasked with generating a question-answer pair based on the text, unconstrained. We then passed this output back into the model with the task of extracting the question and answer components and placed them into a JSON file with the keys \textit{question} and \textit{answer}. We used the same, high temperature of 0.9 for generating the samples and a much lower temperature of 0.2 for extracting into a JSON to reduce the chance of models altering the generated sequences while structuring them.

\section{API-Based LLM Answerability Annotation} \label{app:llm_ans}

To ensure that we evaluate performance on as high-quality questions as possible, we extract only those questions deemed \textit{answerable}, by our definition, by both GPT-4o and Gemini-1.5-pro. Table~\ref{tab:ans_results} shows that the zero-shot samples had the highest rate of predicted answerability; each other variant shows very consistent rates of answerability. This outcome should be tempered by the results in Figure~\ref{fig:error-dist} which indicates that the zero-shot model had an extremely high failure rate in many other regards.

\end{document}